\documentclass{article}

\usepackage[preprint]{neurips_2024}

\usepackage[utf8]{inputenc} 
\usepackage[T1]{fontenc}    
\usepackage{hyperref}       
\usepackage{url}            
\usepackage{booktabs}       
\usepackage{amsfonts}       
\usepackage{nicefrac}       
\usepackage{microtype}      
\usepackage{xcolor}         
\usepackage{amsmath}
\usepackage[inkscapelatex=false]{svg}
\usepackage{caption}
\usepackage{subcaption}
\usepackage{multirow} 
\usepackage{graphicx}

\newcommand{\sysname}{Confucius3-Math}

\newcommand{\reusealg}{Recent Sample Recovery}
\newcommand{\reusealgabbr}{RSR}
\newcommand{\newadv}{Policy-Specific Hardness Weighting}
\newcommand{\newadvabbr}{PSHW}

\title{\sysname: A Lightweight High-Performance Reasoning LLM for Chinese K-12 Mathematics Learning}

\author{%
  Lixin Wu \quad  Na Cai \quad  Qiao Cheng \quad Jiachen Wang \quad Yitao Duan \\
  NetEase Youdao, Beijing, China\\  
  \texttt{\{wulixin, caina, chengqiao, wangjiachen, duan\}@rd.netease.com} \\
}

\begin{document}

\maketitle

\begin{abstract}
We introduce \sysname{}, an open-source large language model with 14B parameters that (1) runs efficiently on a single consumer-grade GPU; (2) achieves SOTA performances on a range of mathematical reasoning tasks, outperforming many models with significantly larger sizes. In particular, as part of our mission to enhancing education and knowledge dissemination with AI, \sysname{} is specifically committed to mathematics learning for Chinese K-12 students and educators. Built via post-training with large-scale reinforcement learning (RL), \sysname{} aligns with national curriculum and excels at solving main-stream Chinese K-12 mathematical problems with low cost. In this report we share our development recipe, the challenges we encounter and the techniques we develop to overcome them. In particular, we introduce three technical innovations: Targeted Entropy Regularization, \reusealg{} and \newadv{}. These innovations encompass a new entropy regularization, a novel data scheduling policy, and an improved group-relative advantage estimator. Collectively, they significantly stabilize the RL training, improve data efficiency, and boost performance. Our work demonstrates the feasibility of building strong reasoning models in a particular domain at low cost. We open-source our model and code at \url{https://github.com/netease-youdao/Confucius3-Math}.

\end{abstract}

\begin{figure}[h]
\centerline{\includegraphics[width=0.85\columnwidth]{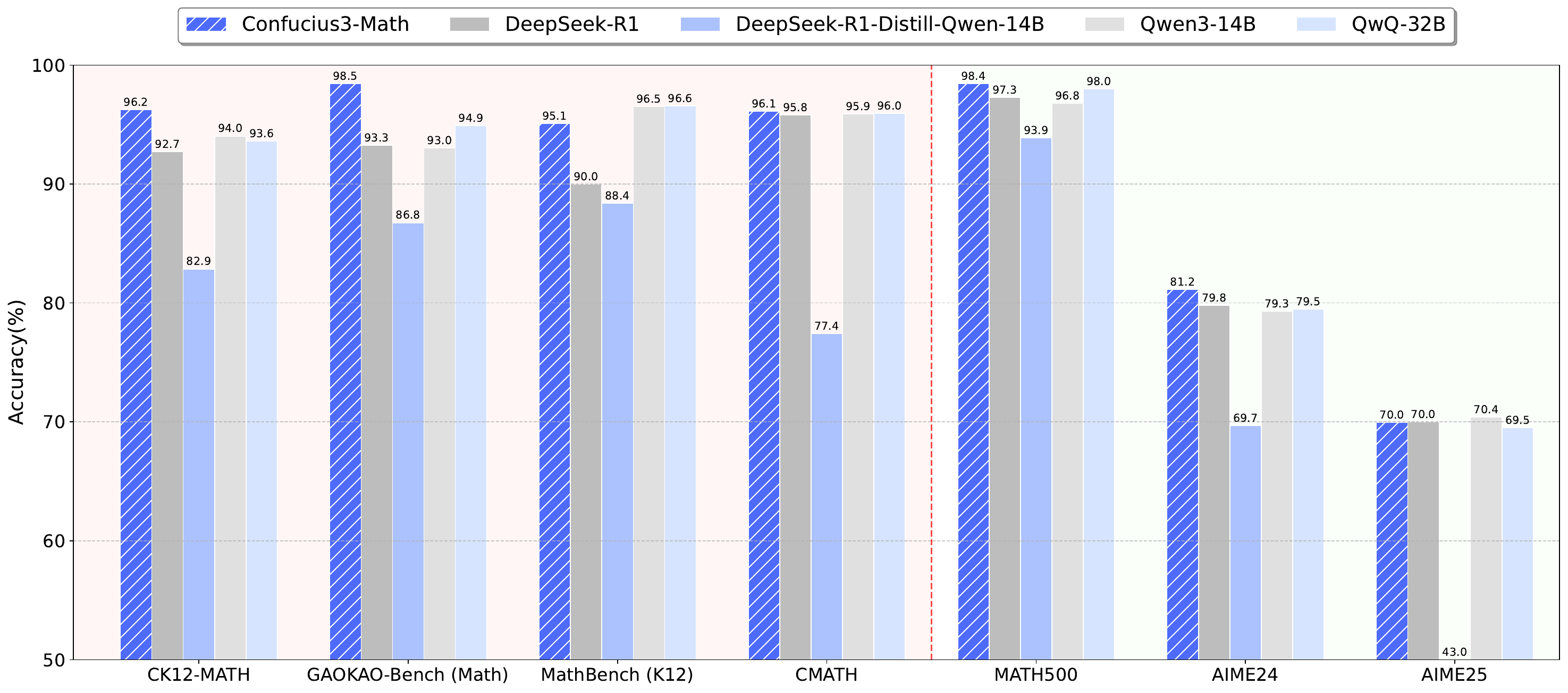}}
  \caption{Benchmark performance of \sysname}
\label{fig.benchmark}
\end{figure}

\section{Introduction}

Reasoning capabilities have been the focus of the latest Large Language Model (LLM) developments. OpenAI's \texttt{o1} model \citep{openai2024o1} represents a major breakthrough in advancing LLM's abilities in solving complex, multi-step tasks such as math and coding, igniting mainstream awareness of reasoning-specific LLMs. \texttt{o1} is known to acquire its performance via Test Time Compute (TTC). However, neither the model weights nor the technical details of \texttt{o1}, or its sequels \cite{openai2025o3-4}, are released. 
TTC is a well-known technique to boost model's performance by allowing the model to ``think'' or compute more at inference time and has been applied successfully in many reasoning tasks, mostly in games such as Backgammon \citep{10.1162/neco.1994.6.2.215}, poker \citep{brown2017safe}, and Go \citep{alphago2016} etc. It takes on many forms from simple consensus/majority vote to beam search and Monte-Carlo Tree Search (MCTS) \citep{Metropolis:1949:MCM}, depending on the application. \citet{jones2021scalingscalinglawsboard} studies scaling laws between train time and test time, clarifying the benefit of TTC, also in the game setting. 

With LLMs that generate results auto-regressively, TTC is more complicated. \citet{brown2024largelanguagemonkeysscaling} shows that inference-time scaling laws still exists with LLMs and the fraction of problems that are solved by any generated sample scales log-linearly with the number of samples\footnote{But the performance plateaus around 100 samples - not a very high ceiling.}. However, this only suggests the \textit{possibility} of TTC's benefits, how to implement a feasible strategy remains challenging. The successes from game AI may not readily transfer to LLM. One major challenge is that the game models are each solving a \textit{single} task (e.g., Go), while the reasoning LLMs are expected to solve potentially infinite number of problem: even the same math problem framed in different words represents a different state for an LLM, resulting in a huge state-action space. Another challenge is that it is much more difficult to verify the correctness of the results.

After the release of \texttt{o1}, there have been many works attempting to replicate it. For example, \citet{o1journey} tries to dissect \texttt{o1}'s mechanisms by detailed analysis of its outputs and focused experiments. Methods exploited include SFT, DPO, tree search etc. \citet{zhao2024marcoo1openreasoningmodels} uses explicit MCTS to boost reasoning performance. These works obtain mild progress and the results are mostly speculations. 

DeepSeek-R1~\citep{deepseekai2025deepseekr1incentivizingreasoningcapability}, released in January 2025, represents a transformative leap in reasoning capabilities and training methodologies. It is the first model to demonstrate that pure reinforcement learning (RL)—without SFT—can effectively cultivate advanced reasoning skills, enabling organic emergent reasoning behaviors like chain-of-thought and self-verification. Technical achievements aside, R1's openness also democratizes access, enriches and strengthens the open LLM ecosystem, paving the way to further advancing reasoning capabilities and enabling practical applications.

\textbf{Reasoning LLM and Education}. We join the endeavor to push the limit of LLM's reasoning capabilities by focusing on a particular domain, namely, K-12 education which has long been regarded as one of the main sectors in which artificial intelligence can bring about disruptive innovations. Education places a much more demanding requirement on model's accuracy, as research has shown that incorrect answers generated by LLMs could have a negative impact on student learning \citep{ali2024chatgpt}. Surprisingly, the most powerful reasoning LLMs that achieve high scores in some complex benchmarks do not perform well on K-12 tasks (see our evaluation in \ref{sec.eval} and studies such as \citet{zhu2024embracingaieducationunderstanding}), calling for model adaptation to this particular domain.

Another major challenge is the inequality caused by the economic disparity of the students. The academic performance of K-12 students has been shown to be highly correlated with their socioeconomic status \citep{cuéllar2024shapingaisimpactbillions}. AI is expected to mitigate the inequality by making educational resources more accessible to all. In particular, LLMs, with their ability to host rich knowledge and support for natural language interaction, represent a promising solution, as they have the potential to provide, to all students with low cost at large-scale, one key element in learning: highly qualified teachers. LLMs have been shown to be effective in improving learning (e.g., \citet{wanggptlearning2025,Kestin__2024}). However, high-performance LLMs are still quite expensive to build and deploy, making their access not affordable for many students from low-income regions. Meanwhile, students who can pay for expensive models have shown a clear advantage in learning~\citep{zhu2024embracingaieducationunderstanding}. Ironically, at the current high cost, the more powerful the LLMs are, the more severe the digital divide becomes.  

We believe that a strong low-cost reasoning LLM goes a long way towards equalizing education. In this work, we release \sysname{} and our experiences in building it. As shown in \ref{fig.benchmark}, \sysname{} is a strong reasoning model achieving SOTA performances on a range of mathematical reasoning tasks and outperforming many models with significantly larger sizes, especially in our target domain, namely Chinese K-12 mathematics learning. The contributions of our work include:

\begin{enumerate}
    \item \sysname{} demonstrates the feasibility of building strong and practical reasoning models in a particular domain at low cost. Specifically, (1) \sysname{} achieves SOTA performance on a number of K-12 math benchmarks; (2) The training of \sysname{} costs only \$26K; and (3) its inference performance is about 15$\times$ that of DeepSeek-R1.
    \item Technically, we show that, given the right base model, a moderate amount of high-quality data, and a good training recipe, post-training via RL is capable of eliciting strong reasoning capabilities in lightweight models.
    \item We introduce Targeted Entropy Regularization, \reusealg{} and \newadv{}. These innovations encompass a new entropy regularization, a novel data scheduling policy, and an improved group-relative advantage estimator. They significantly stabilize the RL training, improve data efficiency, and boost performance.
    \item We open-source \sysname{} and its technical details, inviting the community to utilize it in real-world applications and collaborate on developing practical reasoning LLMs.
    
\end{enumerate}






\section{Approach}

Our goal is to build practical models that enable real-world AI learning applications. To this end, the development of \sysname{} is guided by the following principles:

\begin{enumerate}
    \item Focused Domain: We target the education sector. As a first step towards building LLM models that can effectively assist learning, we especially focus on Chinese K-12 mathematics education.
    
    \item Low Cost: The cost of both developing and deploying the model should be low. We believe that this is crucial for supporting sustainable real-world applications and providing access to all.  
    
    \item High Performance: Research has shown that incorrect answers generated by LLMs could have a negative impact on student learning \citep{ali2024chatgpt}. Therefore, the model must have adequate accuracy to be useful for realistic learning scenarios. 
    
    \item Learning Support: The model should be capable of producing sufficient auxiliary information that are essential part of learning. 
    
    \item Openness: Education is one of the foundations of our social structure with profound and long-term impacts on many people's lives. To ensure its proper development, we believe that its key elements and mechanisms should be open, allowing people from different sectors to scrutinize, collaborate, and contribute.
    
\end{enumerate}


We make the following design choices. First, we choose to build our model upon strong open-source models that already possess powerful capabilities through pre-training and fine-tuning. Recently, post-training has emerged as an effective way to enhance reasoning capabilities and the booming open ecosystem provides great resources for building strong models and applications with low cost. 

Second, we choose model size to be 14B which is large enough to have the expressivity for complex tasks, yet not too large so that, with good engineering, can still run efficiently on consumer-grade GPU.

For post-training, we contemplate a number of choices, including RL, SFT, distillation, and their combinations. For enhancing small models, past works have shown their effectiveness in different situations. For example, DeepSeek-R1~\citep{deepseekai2025deepseekr1incentivizingreasoningcapability} shows that distillation obtains much larger performance gains than post-training the same model via RL. The latest Qwen3 series also use strong-to-weak distillation to optimize their lightweight models (Qwen3-0.6B, 1.7B, 4B, 8B, and 14B) and they show that a combination of off-policy/on-policy distillation achieves significantly better performance than RL~\citep{yang2025qwen3}. 

However, our objective is \textit{not} simply a good model, but a model on par or exceeding even the strongest models with larger sizes in our domain. We believe RL, which encourages emergent behavior, holds greater potential, as indicated by 
its superhuman performance in tasks such as Go~\citep{alphago2016} and protein structure prediction~\citep{alphafold2021}. So we select RL as the main ingredient of our solution. As our experiments turn out, this is quite effective.

\section{Data Curation}

\subsection{Data Sources}
The data used for training the model comes from two major sources: open-source and proprietary.

\textbf{Open-Source Data}. To enhance the model's mathematical capabilities, we collect a large number of open-source English mathematics datasets, including:

\begin{enumerate}
    \item Datasets focused on foundational and advanced mathematics skills, such as GSM8K~\citep{cobbe2021gsm8k}, applied\_math \footnote{\url{https://www.modelscope.cn/datasets/gavinluo/applied\_math.git}.}, and Advanced-Math\footnote{\url{https://www.modelscope.cn/datasets/Haijian/Advanced-Math.git}.}.
    \item Collections of competition-level mathematics problems for various levels, including MATH~\citep{hendrycksmath2021}, NuminaMath-1.5~\citep{numina_math_datasets}.
    \item High-quality reasoning datasets from universities, forums, and research, covering multiple fields such as mathematics, science, puzzles, and coding, including OpenThoughts-114k~\citep{guha2025openthoughtsdatarecipesreasoning}, orz\_math\_57k\_collection~\citep{hu2025openreasonerzeroopensourceapproach}, s1K-1.1~\citep{muennighoff2025s1simpletesttimescaling}, LIMR~\citep{limr2025}, and LIMO~\citep{ye2025limoreasoning}, among others.
\end{enumerate}
The sizes of the datasets are summerarized in table \ref{tab:dataset_samples}.

\begin{table}[htbp]
  \centering
  \begin{tabular}{lr}
    \toprule
    \textbf{Dataset} & \textbf{Number of Samples} \\
    \midrule
    GSM8K Training Set & 7,473 \\
    applied\_math & 10,000 \\
    Advanced-Math & 1,032 \\
    MATH Training Set & 7,497 \\
    NuminaMath-1.5 & 896,215 \\
    OpenThoughts-114k & 113,957 \\
    orz\_math\_57k\_collection & 56879 \\
    s1K-1.1 & 1,000 \\
    LIMR & 1,389 \\
    LIMO & 817 \\
    \hline
    \textbf{Total} & \textbf{1,096,241} \\
    \bottomrule
  \end{tabular}
  \caption{Overview of Open Source Training Datasets}
  \label{tab:dataset_samples}
\end{table}

\textbf{Proprietary Data}. Since we optimize for K-12 mathematics learning scenario, in addition to relevant open-source data, we also collect math questions, and their solutions, accumulated during the operation of our business. They cover various mathematics problems for the domestic K-12 stages (primary, middle, and high school), including a rich variety of types, such as single-choice, multiple-choice, true/false, fill-in-the-blank, calculation, proof, and mixtures of multiple question types etc.

\subsection{Data Preparation}

Prior works \citep{ye2025limoreasoning,kimiteam2025kimik15scalingreinforcement} have shown that the quality and diversity of training data play a crucial role in reinforcement learning. Therefore, we have implemented a rigorous data preprocessing procedure utilizing the NeMo-Curator framework \citep{NeMo}. 

First, we remove test splits from all open-source datasets, as well as all commonly used benchmarks in research. Additionally, since RL requires high-quality data with standard answers, we exclude all synthetic data and data without ground truth.

Next, we execute the following data processing workflow in sequence on open-source data:

\begin{enumerate}
    \item Exact Deduplication: We retain only one instance of identical questions in the dataset.
    \item Fuzzy Deduplication: We deduplicate questions with high Jaccard similarity scores with others.
    \item Semantic Deduplication: We use $k$-means to cluster the embeddings of the questions. Within each cluster, we compute cosine similarities between each pair and remove one question if it is too similar to the other.
    \item Question Type Selection: To get accurate rewards during the RL process, we remove multiple-choice, true/false, and proof questions. The first two types can be guessed and there is no easy way to verify the third.
\end{enumerate}

For our proprietary data, we apply a cleaning stage, as the data originates from mass-scale automated entry with manual correction, it inherently contains significant noise. We do not perform other complex processing since the entering process has avoided duplication and the question types are all friendly to our reward model. 

As a result, we retain approximately 540,000 samples of data for actual training, including 210,000 from open-source data and 330,000 from proprietary data. Note that this is the amount of training data we consume up to this release of \sysname{}. Our data sources, especially proprietary, are producing more. We leave the rest for future improvement.

\section{Base Model Selection}

A good base model with the right initial capabilities could avoid extra cold start, save computation and reach higher performance ceiling for the post-training. Prior to making selection, we conduct extensive explorations on different base models, including Qwen2.5-14B-Base, Qwen2.5-14B-Instruct, DeepSeek-R1-Distill-Qwen-14B, and a previous Confucius reasoning model Confucius-o1-14B. The explorations are conducted via 150 steps of RL. We monitor both the reward scores and policy entropy, which turns out to be a good indicator of the model's exploration ability which is crucial for RL (see discussion in section~\ref{sec.trt}).

We discover that all display certain level of trainability, with DeepSeek-R1-Distill-Qwen-14B performing the best. And indeed, the model's outcome is correlated with its initial entropy. We postulate that the policy entropy could be an importance metric when determining whether a model can serve as an initial model for RL optimization. Interestingly, among the models, Confucius-o1-14B, which is trained from Qwen2.5-14B-Instruct, shows a lower entropy than its base, which appears to be caused by its training method: DPO without entropy regularization.

In addition, we also assess the richness of models' output, as in an education setting, we expect the model to output the problem-solving process in its answer. The reason for emphasizing this soft metric is that in current RL training, the reward module predominantly focuses on the correctness of the response while often overlooking its compliance with the prompt, which largely hinges on the compliance capabilities of its initial model.

We ultimately select DeepSeek-R1-Distill-Qwen-14B model as the base model. Distilled from Qwen2.5-14B using DeepSeek-R1’s data, the model exhibits robust CoT and holds a greater initial edge in the field of mathematics compared to other models of comparable scale. Moreover, its answers align with our expectations for including problem-solving steps.

\section{Reward Modeling}
\label{sec.rm}

We adopt a two-stage, hybrid method combining both rule-based and model-based approaches for reward modeling. 

The first stage acts as a filter, filtering out any output that either does not conform to the right format or contains repetitive text, a problematic phenomenon frequently observed with LLM generation \citep{wang2024mitigatinglanguagemismatchrepetition,10.5555/3666122.3668712}. 
Only the response results that pass these two types of filtering can obtain the subsequent correctness reward score. We adopt this filtering mechanism rather than a reward weighting scheme mainly because the format requirement can be quickly satisfied in the early stage of training. Incorporating it as part of the reward weighting does not have a practical effect in the later stage. In addition, these two types of constraints are also hard requirements. In real usage, we cannot accept a response that does not meet these two types of constraints but is correct in result. 

The second stage rewards the model for its correctness. We use either rule-based or model-based approaches based on different data sources. For open-source data, since the result of each question consists of a single expression, we continue to use conventional rule-based reward modeling. Specifically, we utilize Math-Verify \citep{Math-Verify} for answer extraction and verification. Our proprietary data, on the other hand, contains math problems in forms typically found in Chinese K-12 schools. For example, some problems may contain multiple subquestions, and answers may depend on long textual descriptions far away. Conventional extraction rules often do not work. Therefore, we adopt a model-based approach and use LLM-as-a-Judge to determine the correctness of the model's response. In addition to checking the correctness of the final answer, the reword model also verifies the solving process, as we need it to produce learning materials for students.

Regarding the text repetition filtering, we also find that it is not enough to only scrutinize the answer. Both the thinking and answer parts need to be examined. Otherwise repetition will appear in the thinking part in the early stage of the training and gradually spread to answer region while training progresses, resulting in an irreversible collapse.

\section{Training}

\begin{figure}
\centerline{\includegraphics[width=0.75\columnwidth]{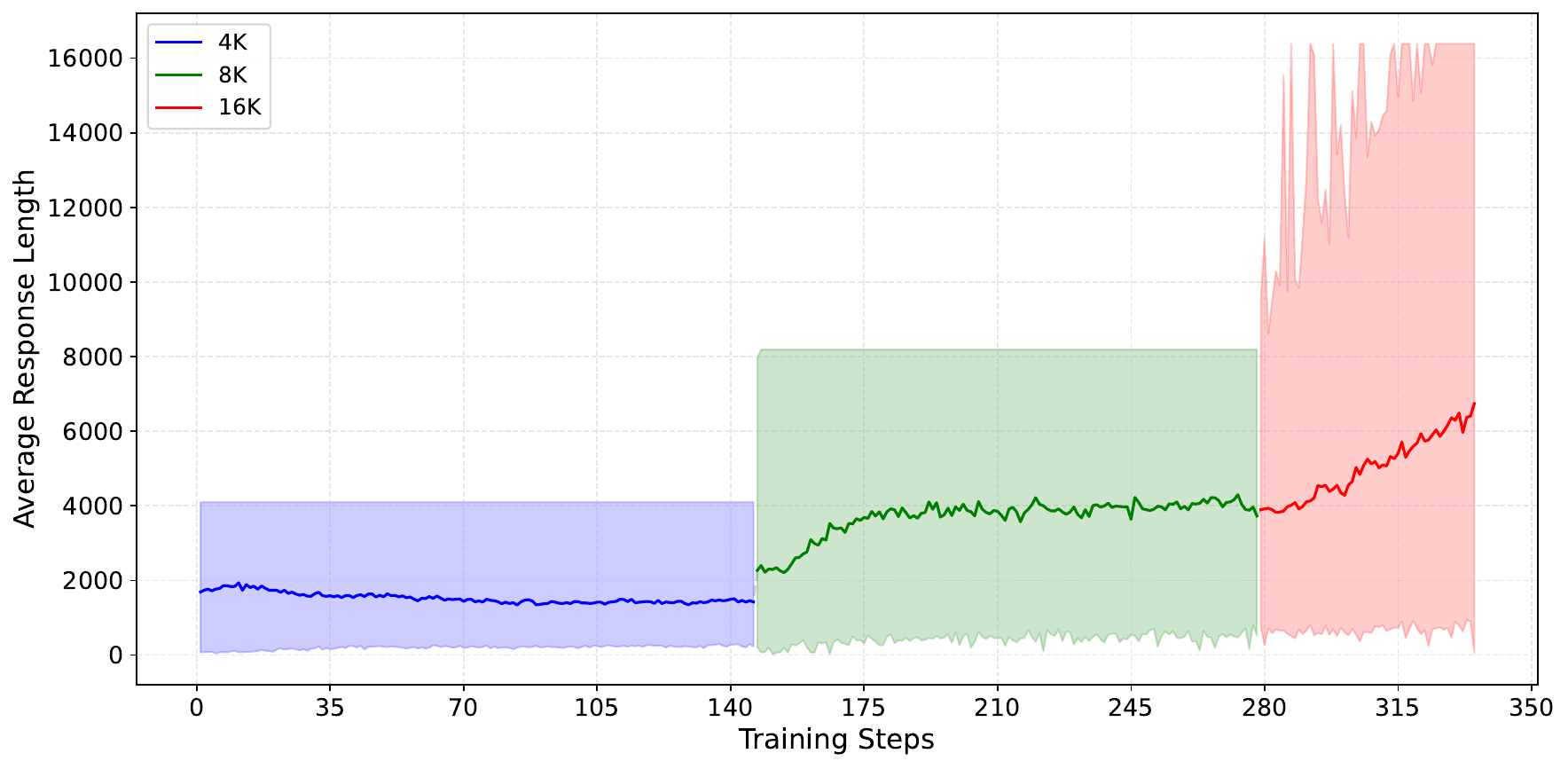}}
\caption{Multi-Stage Length Expansion}
\label{fig.stages}
\end{figure}

We follow a three-stage, pure-RL training pipeline. The staging is mainly for expanding context window, which progressively grows from 4K to 8K, and eventually 16K, while preserving training efficiency. We also adjust training objectives and other configurations as we address some issues we encounter, and with the availability of new tools. We use the same training template as that of R1~\citep{deepseekai2025deepseekr1incentivizingreasoningcapability}
and main-stream optimization frameworks such as Group Relative Policy Optimization (GRPO) \citep{shao2024deepseekmathpushinglimitsmathematical} and Dynamic sAmpling Policy Optimization (DAPO) \citep{yu2025dapoopensourcellmreinforcement}, with our own improvements which will be elaborated in sections \ref{sec.trt}, \ref{sec.pshw}, and \ref{sec.rsr}. A brief summarization of the characteristics of the stages is presented in table \ref{tab.stages}. Figure \ref{fig.stages} shows the changes of average response lengths as the training progresses through the three stages, demonstrating the effectiveness of context expansion and the growth of model's Chain-of-Thoughts (CoT).

\begin{table}[htbp]
  \centering
  \begin{tabular}{lcc}
    \toprule
    \textbf{Stage} & \textbf{Core Algorithms} & \textbf{Context Length} \\
    \midrule
    Stage 1 & GRPO + Targeted Entropy Regularization & 4K \\
    Stage 2 & DAPO +  RSR + PSHW & 8K \\
    Stage 3 & DAPO +  RSR + PSHW & 16K \\
    \bottomrule
  \end{tabular}
  \caption{Training Stage Characteristics}
\label{tab.stages}
\end{table}

In retrospect, we find that our approach, although independently developed, bears two major similarities with concurrent work of Skywork-OR1 \citep{he2025skywork}, namely staged context window expansion and targeted entropy regulation. We elaborate our ideas in subsequent sections.

\subsection{Targeted Entropy Regularization}
\label{sec.trt}

Entropy regularization is a commonly used technique in RL policy optimization. It is believed to benefit exploration by encouraging the selection of more stochastic policies  and preventing the training from prematurely converging on deterministic policies \citep{ahmed2019understanding,snoswell2020revisiting,schulman2017proximal}. It is now commonly used in LLM training (e.g., \citet{yu2025dapoopensourcellmreinforcement}).

However, from our experience, entropy seems to be a delicate matter. LLMs are complex machines with output distribution over a large space of multilingual tokens. The maximum entropy principle appears to have double effects. One major issue we encounter during the training process is the ``mixed language problem'', a phenomenon that also arises with DeepSeek-R1 \citep{deepseekai2025deepseekr1incentivizingreasoningcapability}. This phenomenon appears to be correlated with high entropy, which may encourage the model to substitute tokens with the same meaning from different languages. We tried adding linguistic constraints to the reward, but it was ineffective.

Therefore we introduce ``Targeted Entropy Regularization'' by adding the following term to the loss function

\begin{displaymath}
    |\texttt{entropy\_loss} - \texttt{entropy\_target}| * \texttt{entropy\_coeff}
\end{displaymath}

where \texttt{entropy\_target} is desired entropy level set empirically to 0.55. This is different from the Adaptive Entropy Control in Skywork-OR1 \citep{he2025skywork}: their strategy aims to constrain the lower limit of entropy to prevent entropy collapse, while we constrain both the upper and lower limits. \texttt{entropy\_coeff} is set to 0.001, similar to conventional entropy regularization.


\begin{figure}[h]
\centerline{\includegraphics[width=0.75\columnwidth]{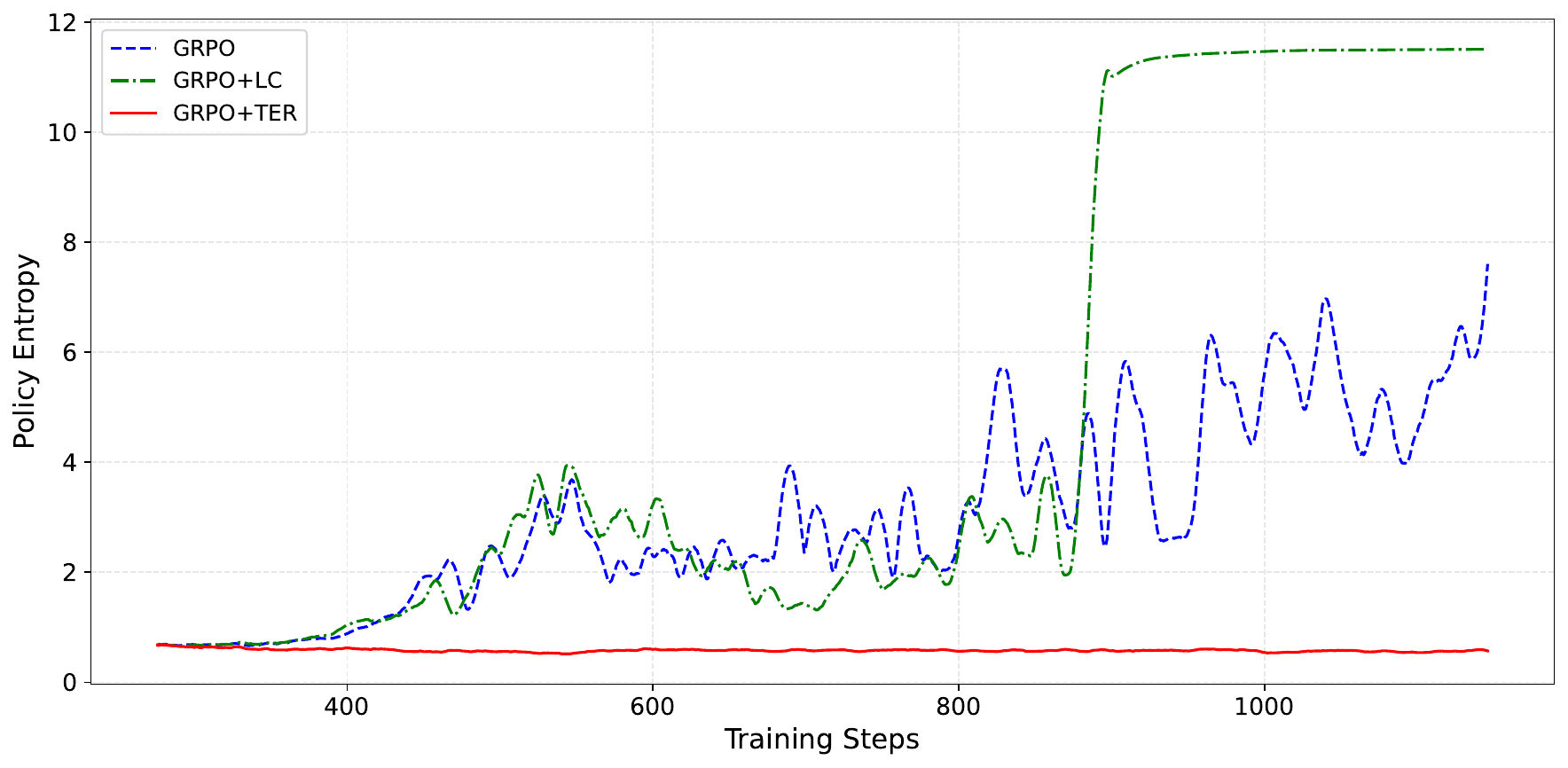}}
\caption{Effects of Target Entropy Regularization on Policy Entropy}
\label{fig.ter}
\end{figure}

Figure \ref{fig.ter} plots the model output entropy using different mechanisms. Clearly Target Entropy Regularization effectively constrains the entropy to a desired level and stabilizes the training. The mixed language problem also goes away. 

Our understanding is that many of the issues are rooted in the stability of training. Once we switch to a more stable training technique, namely our improved DAPO, described in the following sections, after the first stage, those nuances such as KL divergence or entropy regulation are not so useful.

\subsection{\newadv{}}
\label{sec.pshw}

A crucial factor influencing RL fine-tuning is the ability-difficulty gap: the gap between the current policy model's ability to solve the problems and their difficulty levels in the given batch. The gap being both too wide or too narrow could impact the learning efficacy. Traditionally this issue is addressed with curriculum learning: ordering the training data according to their difficulty levels. Curriculum learning has also been applied in LLM RL training. For example,
\citet{wang2025dumpautomateddistributionlevelcurriculum} treats sample selection as an exploration/exploitation problem and models a distribution-level learnability among samples. \cite{chen2025selfevolvingcurriculumllmreasoning} adds a dedicated curriculum policy trained together with the LLM policy. Both adjust sampling probabilities throughout training to get a better sample scheduling. 

We propose a policy specific approach, denoted \newadv{} (\newadvabbr{}). The key insight is, because the policy model is evolving during training, so is the learnability of samples. In other words, learnability should be relative to the current policy. Unlike curriculum learning which schedules training samples, \newadvabbr{} incorporates relative difficulty into advantage estimation which is much simpler and more efficient.

\newadvabbr{} can be used with any compatible RL algorithm including GRPO. We use it with DAPO. Using conventional notation as that of DAPO \citep{yu2025dapoopensourcellmreinforcement}, the algorithm samples a group of outputs $\{o_i\}_{i=1}^G$ for each question $q$ paired with the answer $a$, and optimizes the policy via the following objective:
\begin{equation} \label{eqn.obj}
\begin{aligned}
\mathcal{J}(\theta) =\quad& \mathbb{E}_{(q,a)\sim \mathcal{D}, \{o_i\}_{i=1}^G\sim \pi_{\theta_\text{old}}(\cdot\mid q)}\\&
\Bigg[\sum_{i=1}^{G}\sum_{t=1}^{|o_i|} 
\min \Big( r_{i,t}(\theta) \hat{A}_{i,t},  
\ \text{clip} \Big( r_{i,t}(\theta), 1 - {\varepsilon_{\text{low}}}, 1 + {\varepsilon_{\text{high}}} \Big) \hat{A}_{i,t} \Big) \Bigg] \\
\text{s.t.}\quad& 0< \Big|\{o_i\mid\texttt{is\_equivalent}(a,o_i)\}\Big|< G, 
\end{aligned}
\end{equation}
where
\begin{equation}
    r_{i,t}(\theta)=\frac{\pi_{\theta}(o_{i,t} \mid q, o_{i,<t})}{\pi_{\theta_{\text{old}}}(o_{i,t} \mid q,o_{i,<t})},\quad\hat{A}_{i,t} = \left[{R_i - \mu}\right]D(q), \quad \mu = \text{mean}(\{R_i\}_{i=1}^G)
\end{equation}

and

\begin{equation}
    R_i = 
    \begin{cases} 
    1, & \texttt{is\_equivalent}(a, o_i) \\
    -1, & \text{otherwise} 
    \end{cases}, \quad
    D(q) = \alpha \times \mu + 1.256.
\end{equation}

We set the coefficient $\alpha = -0.256$ empirically. In essence, the mean score $\mu$ represents how well the policy does on question $q$, which we believe is a good measure of the difficulty of $q$ \textit{relative} to the current policy. \newadvabbr{} implements a kind of adaptive difficulty-aware advantage estimator which encourages the policy to optimize and explore problems it perceives as ``hard''.

We also introduce another modification to DAPO, as obvious from equation \ref{eqn.obj}. Namely, as observed by \citet{liu2025understandingr1zeroliketrainingcritical}, length normalization used in GRPO, and DAPO as well, tends to produce a response length bias. For correct responses, the algorithm favors brevity in correct answers; conversely, the normalization reduces the penalty for longer incorrect responses, which leads the policy to prefer longer incorrect responses. Therefore, similar to \citet{liu2025understandingr1zeroliketrainingcritical}, we remove the length normalization.

\subsection{\reusealg{}}
\label{sec.rsr}

One of the innovations introduced by DAPO \citep{yu2025dapoopensourcellmreinforcement} that prove to be quite effective in boosting training efficiency is Dynamic Sampling: for each batch, DAPO over-samples and filters out prompts with accuracy equal to 1 or 0. This ensures that the policy is updated with effective samples.

However, in order to fill each batch, over-sampling results in generating much more samples that are wasted. We observe up to 3 to 9 times more samples generated for each batch. Among them, some are filtered out. Still, a significant portion of the the remaining legitimate samples are also discarded because of the batch size constraint.

Ostensibly, it does not make sense to recover these samples for future use because they are generated from an old policy. Also the situation is different from experience replay \citep{er-lin92} which reuses old experiences in an \textit{off-policy} setting. Here we are doing \textit{on-policy} learning and the samples are never used. Nevertheless, we experiment saving those legitimate overflow samples and using them in the next batch, a method we denote \reusealg{} (\reusealgabbr{}). 

An orthodox way should have incorporated importance sampling in the process to account for the disparity in distributions. However, our experiments do not produce much difference so for simplicity we do not use it. Detailed algorithm is presented in algorithm \ref{algo:dapo}:

\vspace{-3pt}
\setcounter{table}{0}
\begin{table}[h]
    \centering
    \begin{tabular}{@{}p{1.0\textwidth}@{}} 
        \toprule 
        \textbf{Algorithm 1} \; DAPO+\textbf{RSR}: DAPO with \textbf{R}ecent \textbf{S}ample \textbf{R}ecovery \\
        \midrule 
        \textbf{Input} initial policy model $\pi_\theta$; reawrd model $R$; task prompts $\mathcal{D}$; \textcolor{red}{batch size $N$}; hyperparameters $\varepsilon_\mathtt{low}, \varepsilon_\mathtt{high}$ \\
        \;1: \; \textcolor{red}{Initialize recovery buffer $B_{r}=\emptyset$} \\
        \;2: \; \textbf{for} step = 1,...,M \textbf{do} \\
        \;3: \;\;\;\;\; \textcolor{red}{if $|B_{r}| > 0$:} \\
        \;4: \;\;\;\;\;\;\;\;\; \textcolor{red}{Add $B_{r}$ to dynamic sampling buffer $B_{c}$} \\
        \;5: \;\;\;\;\;\;\;\;\; \textcolor{red}{Set $B_{r}=\emptyset$} \\
        \;6: \;\;\;\;\; Sample a batch $\mathcal{D}_b$ from $\mathcal{D}$ \\
        \;7: \;\;\;\;\; Update the old policy model $\pi_{\theta_{old}} \leftarrow \pi_\theta$\\
        \;8: \;\;\;\;\; Sample \textit{G} outputs $\{o_i\}_{i=1}^{G} \sim \pi_{\theta_{\text{old}}}(\cdot | q)$ for each question $q \in \mathcal{D}_b$ \\
        \;9: \;\;\;\;\; Compute rewards $\{r_i\}_{i=1}^{G}$ for each sampled output $o_i$ by running $R$ \\
        \;10: \;\;\; Filter out $o_i$ and add the remaining to the dynamic sampling buffer $B_{c}$ \\
        \;11: \;\;\; \textbf{if} $|B_{c}| <N$: \\
        \;12: \;\;\;\;\;\;\; \textbf{continue} \\
        \;13: \;\;\; \textbf{else}: \\
        \;14: \;\;\;\;\;\;\; \textcolor{red}{Save remaining legitimate samples for subsequent use as $B_{r}=B_{c}[N:]$} \\
        \;15: \;\;\;\;\;\;\; Select the currently used data as $B_{c}=B_{c}[:N]$ \\
        \;16: \;\;\; For each $o_i$ in the $B_{c}$, compute $\hat{A}_{i,t}$ for the \textit{t}-th token of $o_i$  \\
        \;17: \;\;\; \textbf{for} iteration = 1, ..., $\mu$ \textbf{do}\\
        \;18: \;\;\;\;\;\;\; Update the policy model $\pi_\theta$ by maximizing the DAPO objective \\
        \textbf{Output} $\pi_\theta$\\
        \bottomrule
    \end{tabular}
    \captionsetup{labelformat=empty}
    \caption{}
    \label{algo:dapo}
\end{table}
\vspace{-12pt}

\begin{figure}[h]
\centering
\begin{subfigure}[b]{0.75\textwidth}
\centerline{\includegraphics[width=\columnwidth]{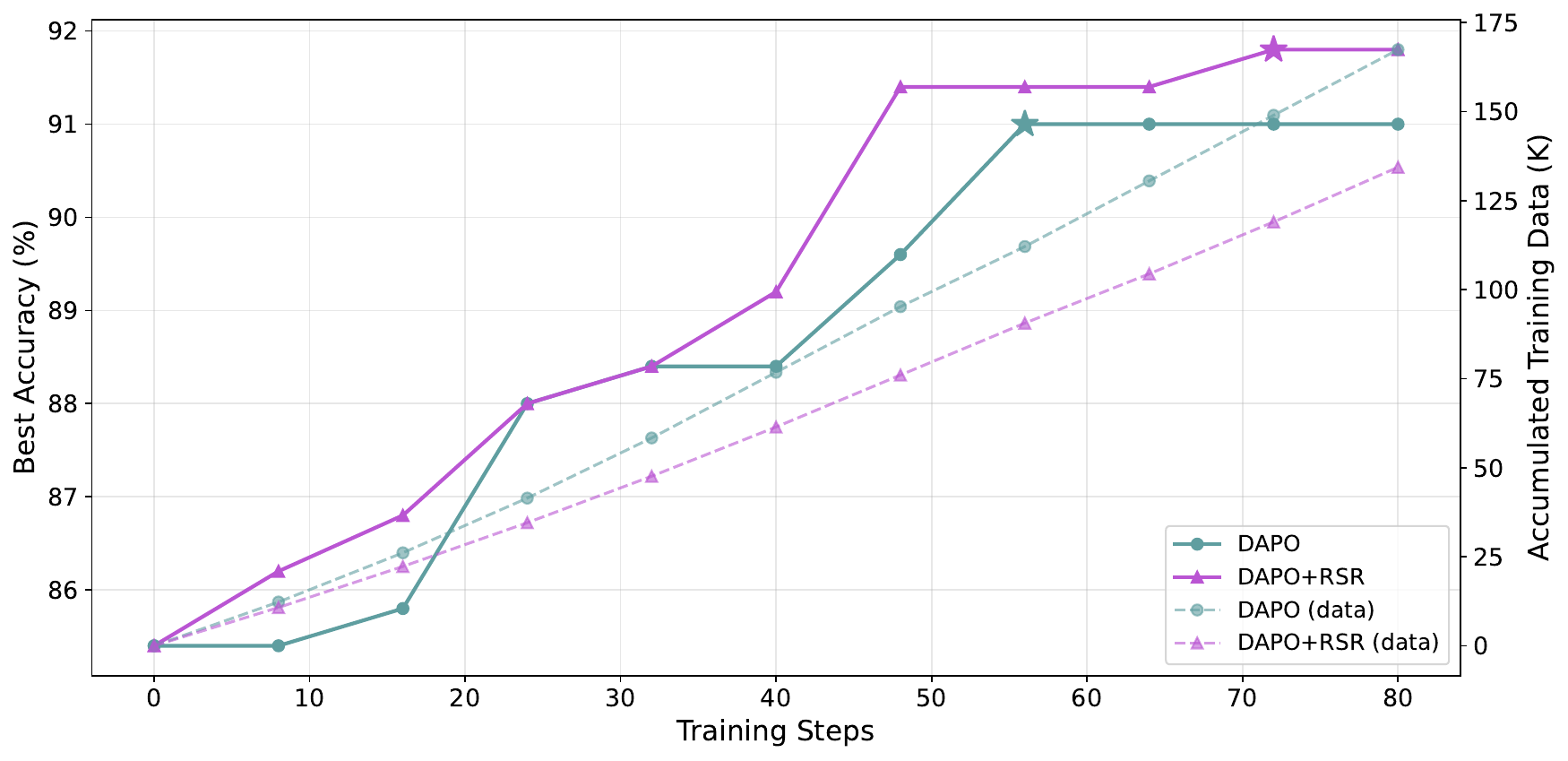}}
\caption{Training Efficiency}
\label{fig.rsr_train_efficiency}
\end{subfigure}
\begin{subfigure}[b]{0.75\textwidth}
\centerline{\includegraphics[width=\columnwidth]{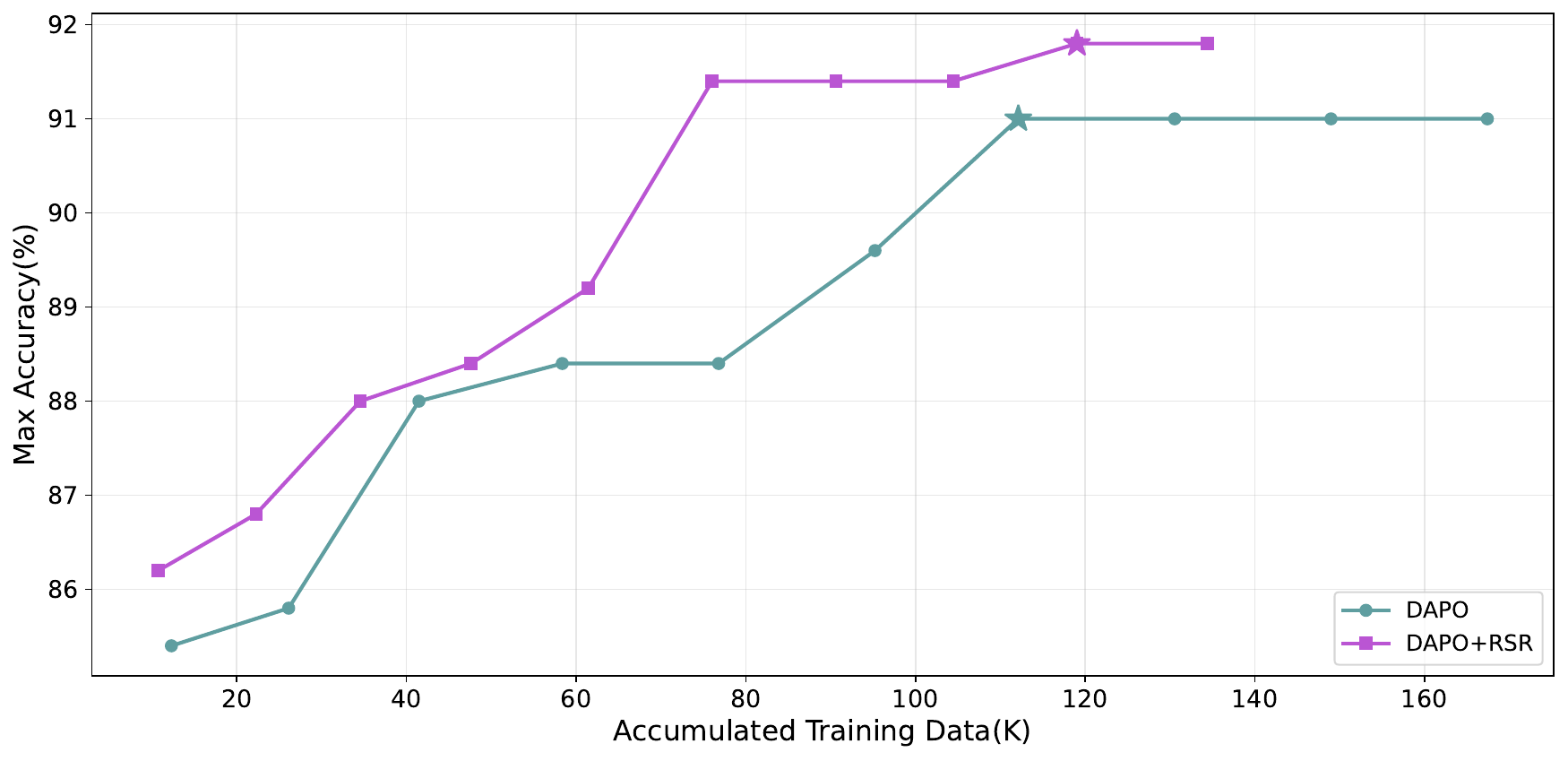}}
\caption{Data Efficiency}
\label{fig.rsr_data_efficiency}
\end{subfigure}

\caption{RSR+DAPO vs. DAPO}
\label{fig.rsr}
\end{figure}

Figures \ref{fig.rsr} show the effects of \reusealgabbr{} on training quality and data efficiency. Figure \ref{fig.rsr_train_efficiency} shows the optimal model quality on evaluation set (solid line, left vertical axis) and the actual amount of raw training data consumed (dashed line, right vertical axis) during the training process. The $x$-axis is the number of training steps, which is proportional to the actual effective samples consumed since each batch consists of a constant number of 256 samples. The solid purple line is always above the solid green line, indicating \reusealgabbr+DAPO's superior training performance. The dashed purple line, on the other hand, grows with a slower rate than the dashed green line. At the same training steps, the \reusealgabbr{} strategy consumes less training data, indicating that it can improve the utilization of training data. Figure \ref{fig.rsr_data_efficiency} on the right figure is a more intuitive demonstration of \reusealgabbr{}'s data efficiency. With the same amount of raw training data, \reusealgabbr{} always outperforms the non-\reusealgabbr{} strategy, with a quality difference reaching up to a 3 percent points at around the 80k data point.

Interestingly, \reusealgabbr{} not only improves data efficiency, but also produces better results, even with the same amount of \textit{effective} data, as indicated by the two solid lines in figure \ref{fig.rsr_train_efficiency}. We speculate that this may be due to certain smoothing effect, since \reusealgabbr{} blends two distributions that are not too far away. We will give this issue a more rigorious treatment in another paper.



\section{Evaluation}

\label{sec.eval}

\textbf{Benchmarks}. We evaluate \sysname{} on several benchmarks relevant to our target domain. CK12-MATH (Chinese K-12 Math) is our internal benchmark sampled from \textit{actual} user queries. It contains 500 math problems used in daily learning, such as homeworks and quizs, in typical Chinese schools. The performance on this dataset represents the actual experiences of a typical Chinese K-12 student interacting with the models. We also select several commonly used public benchmarks for quality assessment. These include: the math subset of GAOKAO-Bench \citep{zhang2023evaluating}, which contains problems from China’s National College Entrance Exam (GAOKAO), the K12 subset in MathBench \citep{liu2024mathbench}, and CMATH \citep{wei2023cmath}, a Chinese Elementary School Math Word Problems dataset. In addition, we also include some widely used competition-level math datasets: MATH500, AIME24, and AIME25.

\textbf{Baselines}. We compare against DeepSeek-R1, the SOTA open-source reasoning model with 671B parameters, our base model DeepSeek-R1-Distill-Qwen-14B, as well as two other strong open-source models: Qwen3-14B~\citep{yang2025qwen3} and QwQ-32B~\citep{QwQ-32B}. Qwen3-14B is one of the latest Qwen3 series with size comparable to \sysname{}. Pre-trained using approximately 36 trillion tokens, nearly twice the amount of Qwen2 (and our base model), and then fine-tuned with a combination of off-policy/on-policy distillation using data from Qwen3 flagship models~\citep{yang2025qwen3}, Qwen3-14B stands for the current SOTA model at its scale. QwQ-32B~\citep{QwQ-32B} is also a strong reasoning model but with over twice the size of \sysname{}. These baselines represent different design choices in the landscape of building practical domain models.

\textbf{Evaluation Setup}. We extract corresponding performance data from their technical report if available, and use the following inference setting if not. For each model, we use the official system prompt, except for R1, for which we find that not using the system prompt leads to higher quality. The maximum generated response tokens is set to 32,768 for all models. To avoid the text repetition problem with greedy decoding found in later RL training stage of DeepSeek and Qwen3 \citep{deepseekai2025deepseekr1incentivizingreasoningcapability,yang2025qwen3}, we use the sampling strategy to generate $k$ response results and reported the pass@1 results. Specifically, for our model, we use a sampling temperature of 1.0 and a top-$p$ value of 0.7, while for other models, we use the sampling parameters recommended by the official documentation. The value of $k$ is set differently for different benchmarks for comparability and stability. For MATH500, AIME24, and AIME25, we follow DeepSeek's practice of setting $k=64$, and for other test sets, $k$ is set to approximately $2,000/N$ where $N$ is the number of samples in the set. Pass@1 is then calculated as$$pass@1=\frac{1}{k}\sum_{i=1}^{k}p_{i}$$ where $p_{i}$ denotes the correctness of the $i$-th response.

Additionally, it is worth mentioning that our internal CK12-MATH benchmark also contains the intermediate steps in a problem's solution so we use a more stringent standard for verifying \sysname{} since it is trained to generate these steps in the answer block. Namely, we also evaluate the correctness of these steps using LLM-as-a-judge. A problem is deemed solved by \sysname{} only when both the final answer and the intermediate steps are correct.

\textbf{Results}. The main results are presented in table \ref{tab.benchmark}. It shows the accuracy rates of the models on a number of benchmarks. The same data is also shown visually in figure \ref{fig.benchmark} in the beginning of this paper. The first 4 rows in table \ref{tab.benchmark}, as well as the left part of figure \ref{fig.benchmark} with pink background, are the benchmarks pertinent to general K-12 learning, while the rest are competition-level benchmarks. \sysname{} outperforms all other models on all but two benchmarks, on which \sysname{}'s performance is comparable to the strongest models. In particular, on the CK12-MATH benchmark, which is the most important focus of our application scenario, \sysname{} leads DeepSeek-R1 by 3.5 points. Qwen3-14B, as a 14B model, also performs quite well on some tasks. However, Qwen3-14B starts with a much stronger initial model, consumes more data, and relies on more powerful teacher models. We believe \sysname{}'s RL solution points to an alternative, and maybe more effective, technical route. As mentioned earlier, we still have a large amount of unused data relevant to this scenario and continued improvement is almost certain. This high accuracy, plus its low inference cost (see section \ref{sec.inference}), makes \sysname{} a viable LLM solution in our actual business operations.  

We also list, in parentheses, the score improvements of \sysname{} against its base, DeepSeek-R1-Distill-Qwen-14B, in the last column of table \ref{tab.benchmark}. \sysname{} obtains large gains across all benchmarks, with the largest reaching 26.98 points. They demonstrate the effectiveness of our RL optimization process. 

\begin{table}[h]
    \setcounter{table}{2}
    \centering
    \footnotesize
    \setlength{\tabcolsep}{1.9pt}
    \begin{tabular}{@{}c | l | c  c  c | c |c c@{}}
    \toprule
    & \multirow{2}{*}{\centering \textbf{\,\,\,Benchmark {\tiny (subset)}}}  & \multirow{2}{*}{\textbf{DeepSeek-R1}}  & \multirow{2}{*}{\textbf{Qwen3-14B}} & \multirow{2}{*}{\textbf{QwQ-32B}} & \textbf{DeepSeek-R1-} & \multirow{2}{*}{\textbf{\sysname}} \\
    & & \textbf{}  & \textbf{} & \textbf{} & \textbf{Distill-Qwen-14B}& \textbf{} \\
    \midrule
    \multirow{4}{*}{K-12} & CK12-MATH & 92.74 & 94.04 & 93.60 & 82.86 & \textbf{96.24 (+13.38)} \\
    & GAOKAO-Bench \tiny{(Math)} & 93.27 & 93.04 & 94.93 & 86.75 & \textbf{98.46 (+11.71)} \\
    & MathBench \tiny{(K12)} & 89.99 & 96.51 & \textbf{96.57} & 88.40 & 95.10 (+6.70)\\
    & CMATH & 95.81 & 95.90 & 95.95 & 77.41 & \textbf{96.13 (+18.72)} \\
    \midrule
    \midrule

    \multirow{3}{*}{Comp} & MATH-500 & 97.30* & 96.80* & 98.00* & 93.90* & \textbf{98.44 (+4.54)} \\
    & AIME 2024 & 79.80* & 79.30* & 79.50* & 69.70* & \textbf{81.15 (+11.45)} \\
    & AIME 2025 & 70.00* & \textbf{70.40*} & 69.50* & 42.97 & 69.95 (+26.98) \\
    \bottomrule
    \end{tabular}
    \caption{Comparison between \sysname{} and other representative models. Items with asterisk are taken from their own publications. Bold items are the best performing among all models tested. The numbers in parentheses in the last column are score lifts of \sysname{} against its base, DeepSeek-R1-Distill-Qwen-14B.
    }
    \label{tab.benchmark}
\end{table}

\section{Inference Performance}
\label{sec.inference}

The \sysname{} model outperforms DeepSeek-R1 671B on K-12 math reasoning tasks. With only 14B parameters, it can be deployed with much lower cost, a crucial factor affecting the sustainability of a service. We conduct tests to verify \sysname{}'s inference efficiency under two configurations: (1) general throughput measurements with abundant resources, and (2) low-resource deployment feasibility test. For (1), we run both R1 and \sysname{} on the same high-performance hardware and measure their throughput, in terms of tokens per second. Specifically, we use a machine with the following hardware components:
\begin{itemize}
\item 8 $\times$ NVIDIA H800 SXM5 GPUs with 80 GB HBM3 memory
\item 112 Intel(R) Xeon(R) Gold 6330 CPU @ 2.00GHz
\end{itemize}

The models are deployed using the vLLM 0.9.0 framework, and CUDA (version 12.8). Both prefix caching and chunk filling are enabled during these tests. We choose FP8 as the model quantization level. For DeepSeek-R1, a parallel strategy with TP = 8 is employed.  \sysname{} inference is run on a single GPU while DeepSeek-R1 671B on 8, to make sure R1 is deployed with adequate resources. We scale the results of \sysname{} by a factor of 8 for the two to be comparable. All other parameters are configured to the default settings of the vLLM framework.

A model with low resource requirement is advantageous, as it has more application opportunities. For this test we deploy \sysname{} with FP8 precision on a single GeForce RTX 4090D GPU with 24GB memory. Again we scale the numbers by 8 for comparability. Note that it is not possible to run DeepSeek-R1 in this setting at all. 

The test data are sampled from user queries with an average input length of 115 tokens. We measure throughput for both input and output under different request rates (QPS).

\begin{figure}[htbp]
  \centering
  \includegraphics[width=0.75\textwidth]{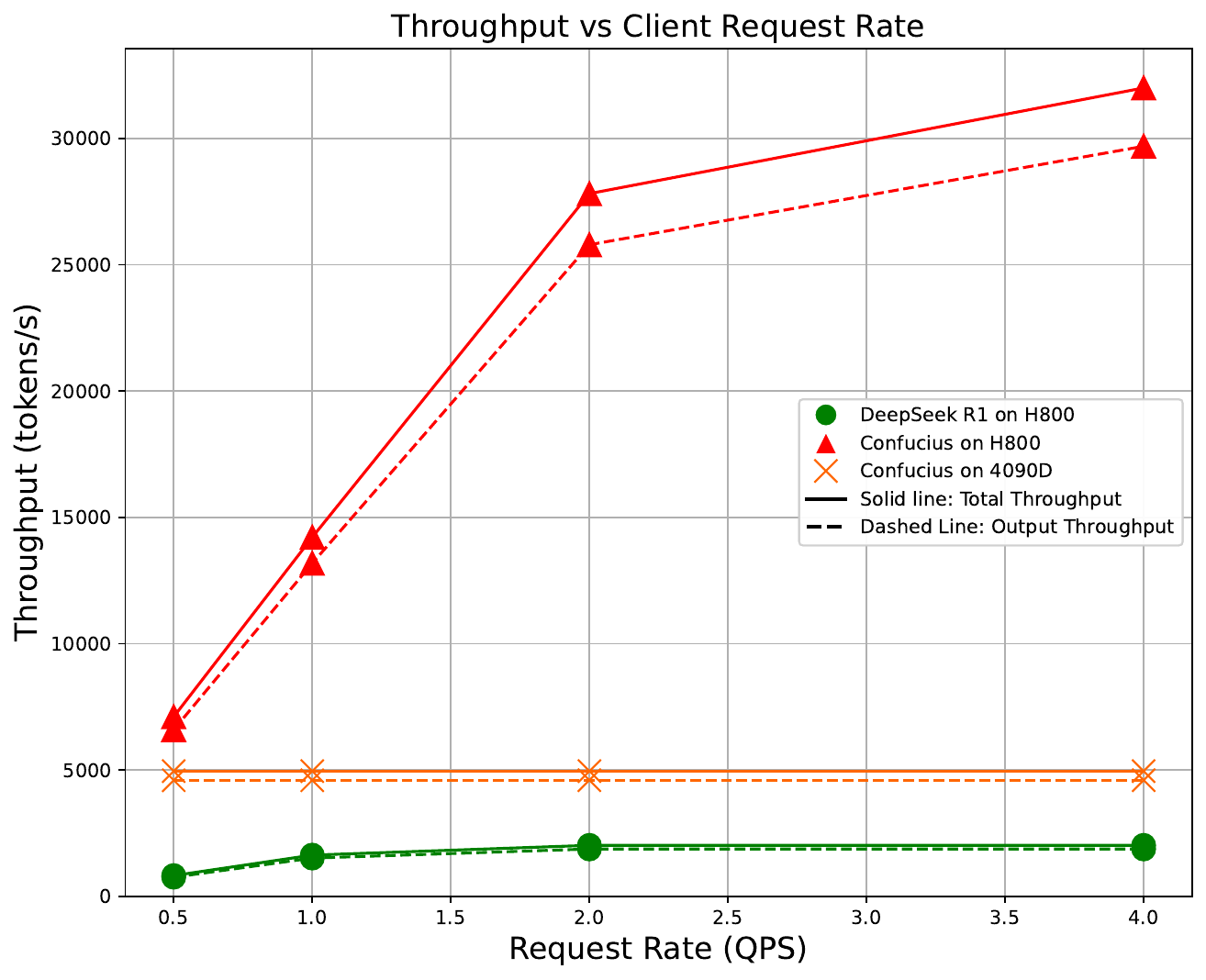}
  \caption{Deployment Performance Result}
  \label{fig:dpr}
\end{figure}

Results are shown in figure \ref{fig:dpr}. For each run, solid line represents total throughput, in tokens per second, while dashed line is output throughput, also in tokens per second. DeepSeek-R1 and \sysname{}'s performances on H800 are shown as green dots and red triangles, respectively. \sysname{}'s throughputs on 4090D are represented as red $\times$.

A few patterns are obvious. Firstly, \sysname{} outperforms R1 by large margins at all request rates. Secondly, R1 saturates quickly at 1,513 and 1,631 tokens per second on output and total throughput, respectively, at QPS = 1. This means that the system reaches its processing capacity at very low workload. \sysname{}, on the other hand, starts with 13,182 and 14,211 tokens per second at QPS = 1, scales more gracefully as QPS grows. It reaches its peak throughput at QPS 4, with a total throughput of 31,994. Compared with DeepSeek-R1, this is a 15.8$\times$ speedup.

Thirdly, even running on low-resource hardware, \sysname{} surpasses DeepSeek-R1's performance on H800. The model achieves impressive inference input and output throughputs of up to 4,596 and 4,956 tokens per second, respectively, more than double those of R1.

\section{Feasibility and Cost}

An alternative solution, strong-to-weak distillation, is a commonly used method to enhance small models. It is actually more complex than it appears and may involve hidden costs. Acquiring outputs from teacher models can be done via API for proprietary models or hosting inference services for open models. Neither is free. And the more powerful the teacher, the more expensive. In particular, on-policy distillation, such as that used in training Qwen3-14B~\citep{yang2025qwen3}, requires more frequent and real-time access to the teacher model. This is only possible when one has access to the model weights, further narrowing down one's choices. Again, powerful models are usually large and hosting them is not a negligible cost. For example, on-policy distillation using DeepSeek-R1 as the teacher model needs, in addition to training GPUs, at least one extra H800-like server (to host the teacher) to run, let alone efficiently.

In contrast, our post-training pipeline is simple, consisting of only RL stages. We can train our model using a \textit{single} H800 server (8 GPUs). This is a much lower investment requirement. 

\begin{table}[h]
\centering
\begin{tabular}{l|ccc|c}
\toprule
\textbf{Training Costs} & \textbf{Stage 1} & \textbf{Stage 2} & \textbf{Stage 3} & \textbf{Total} \\
\midrule
in H800 GPU Hours & 4,234 & 5,470 & 3,405 & 13,109 \\
in USD & \$8.5K & \$10.9K & \$6.8K & \$26K \\
\bottomrule
\end{tabular}
\caption{Training costs of \sysname{}, assuming the rental price of H800 is \$2 per GPU hour.}
\label{tab:training_costs}
\end{table}

The training cost of \sysname{}, divided into stages, is presented in table \ref{tab:training_costs}. The entire training process takes 13,109 GPU hours and costs \$26K.

The inference cost is also very low. Under full-load operation, a single H800 server can process 106.9 million tokens per hour. This translates into \$0.15 per million tokens, far more affordable than most general-purpose LLMs.

\section{Conclusions and Future Work}

\sysname{} demonstrates the feasibility of building strong reasoning models in a particular domain at low cost. In particular, it shows that, given the right base model, a moderate amount of high-quality data, and a good training recipe, RL is capable of producing strong reasoning capabilities in small models.

We believe that the potential of RL has not been fully exploited. As we mentioned earlier, \sysname{} only consumes a small portion of our data. We expect the model performance to continue to improve as we carry on with the training.

There are also many other promising directions that we are exploring. LLM post-training provides great application opportunities for RL, as well as many challenges. For example, the targeted entropy regulation method used in this work is effective. However, it lacks a principled treatment. Fundamentally, the issue it addresses is the exploration–exploitation dilemma, a problem extensively studied in conventional RL. We believe it is worth investigating the issue using mature frameworks such as Bayesian multi-armed bandit~\citep{scott2010Bayesianbandit}.

Application-wise, we have been focusing on mathematical capabilities so far and the model is good at solving math problems. Real-world education and learning scenarios call for more comprehensive functions such as supporting more subjects (e.g., language learning), academic evaluation, homework correction, personalized learning etc. Confucius3 series models will incorporate them in the near future. We believe that conditions are ripe for LLMs to make a real impact on education and we will continue to push their capacity to support our mission to equalizing education with AI.

\begin{ack}
We thank the open-source community, especially Qwen, DeepSeek, and verl, among others, for their generous sharing of models, technical expertise, and tools that have empowered us in our endeavors. We are inspired by their selfless acts of collaboration and are committed to sharing our own findings and resources to support others in their journeys and advance AI boundaries.

\end{ack}

\bibliographystyle{icml2024}

\bibliography{confucius}

\begin{thebibliography}{46}
\providecommand{\natexlab}[1]{#1}
\providecommand{\url}[1]{\texttt{#1}}
\expandafter\ifx\csname urlstyle\endcsname\relax
  \providecommand{\doi}[1]{doi: #1}\else
  \providecommand{\doi}{doi: \begingroup \urlstyle{rm}\Url}\fi

\bibitem[Ahmed et~al.(2019)Ahmed, Le~Roux, Norouzi, and Schuurmans]{ahmed2019understanding}
Ahmed, Z., Le~Roux, N., Norouzi, M., and Schuurmans, D.
\newblock Understanding the impact of entropy on policy optimization.
\newblock In \emph{International conference on machine learning}, pp.\  151--160. PMLR, 2019.

\bibitem[Ali et~al.(2024)Ali, Fatemi, Boskabadi, Nikfar, Ugwuoke, and Ali]{ali2024chatgpt}
Ali, D., Fatemi, Y., Boskabadi, E., Nikfar, M., Ugwuoke, J., and Ali, H.
\newblock Chatgpt in teaching and learning: A systematic review.
\newblock \emph{Education sciences}, 14\penalty0 (6):\penalty0 643, 2024.

\bibitem[Brown et~al.(2024)Brown, Juravsky, Ehrlich, Clark, Le, Ré, and Mirhoseini]{brown2024largelanguagemonkeysscaling}
Brown, B., Juravsky, J., Ehrlich, R., Clark, R., Le, Q.~V., Ré, C., and Mirhoseini, A.
\newblock Large language monkeys: Scaling inference compute with repeated sampling, 2024.
\newblock URL \url{https://arxiv.org/abs/2407.21787}.

\bibitem[Brown \& Sandholm(2017)Brown and Sandholm]{brown2017safe}
Brown, N. and Sandholm, T.
\newblock Safe and nested subgame solving for imperfect-information games.
\newblock \emph{Advances in neural information processing systems}, 30, 2017.

\bibitem[Chen et~al.(2025)Chen, Lu, Kim, Zhang, Tang, Piché, Gontier, Bengio, and Kamalloo]{chen2025selfevolvingcurriculumllmreasoning}
Chen, X., Lu, J., Kim, M., Zhang, D., Tang, J., Piché, A., Gontier, N., Bengio, Y., and Kamalloo, E.
\newblock Self-evolving curriculum for llm reasoning, 2025.
\newblock URL \url{https://arxiv.org/abs/2505.14970}.

\bibitem[Cobbe et~al.(2021)Cobbe, Kosaraju, Bavarian, Chen, Jun, Kaiser, Plappert, Tworek, Hilton, Nakano, Hesse, and Schulman]{cobbe2021gsm8k}
Cobbe, K., Kosaraju, V., Bavarian, M., Chen, M., Jun, H., Kaiser, L., Plappert, M., Tworek, J., Hilton, J., Nakano, R., Hesse, C., and Schulman, J.
\newblock Training verifiers to solve math word problems.
\newblock \emph{arXiv preprint arXiv:2110.14168}, 2021.

\bibitem[Cuéllar et~al.(2024)Cuéllar, Dean, Doshi-Velez, Hennessy, Konwinski, Koyejo, Moiloa, Pierson, and Patterson]{cuéllar2024shapingaisimpactbillions}
Cuéllar, M.-F., Dean, J., Doshi-Velez, F., Hennessy, J., Konwinski, A., Koyejo, S., Moiloa, P., Pierson, E., and Patterson, D.
\newblock Shaping ai's impact on billions of lives, 2024.
\newblock URL \url{https://arxiv.org/abs/2412.02730}.

\bibitem[DeepSeek-AI et~al.(2025)DeepSeek-AI, Guo, Yang, Zhang, Song, Zhang, Xu, Zhu, Ma, Wang, Bi, Zhang, Yu, Wu, Wu, Gou, Shao, Li, Gao, Liu, Xue, Wang, Wu, Feng, Lu, Zhao, Deng, Zhang, Ruan, Dai, Chen, Ji, Li, Lin, Dai, Luo, Hao, Chen, Li, Zhang, Bao, Xu, Wang, Ding, Xin, Gao, Qu, Li, Guo, Li, Wang, Chen, Yuan, Qiu, Li, Cai, Ni, Liang, Chen, Dong, Hu, Gao, Guan, Huang, Yu, Wang, Zhang, Zhao, Wang, Zhang, Xu, Xia, Zhang, Zhang, Tang, Li, Wang, Li, Tian, Huang, Zhang, Wang, Chen, Du, Ge, Zhang, Pan, Wang, Chen, Jin, Chen, Lu, Zhou, Chen, Ye, Wang, Yu, Zhou, Pan, Li, Zhou, Wu, Ye, Yun, Pei, Sun, Wang, Zeng, Zhao, Liu, Liang, Gao, Yu, Zhang, Xiao, An, Liu, Wang, Chen, Nie, Cheng, Liu, Xie, Liu, Yang, Li, Su, Lin, Li, Jin, Shen, Chen, Sun, Wang, Song, Zhou, Wang, Shan, Li, Wang, Wei, Zhang, Xu, Li, Zhao, Sun, Wang, Yu, Zhang, Shi, Xiong, He, Piao, Wang, Tan, Ma, Liu, Guo, Ou, Wang, Gong, Zou, He, Xiong, Luo, You, Liu, Zhou, Zhu, Xu, Huang, Li, Zheng, Zhu, Ma, Tang, Zha, Yan, Ren, Ren, Sha, Fu, Xu, Xie, Zhang,
  Hao, Ma, Yan, Wu, Gu, Zhu, Liu, Li, Xie, Song, Pan, Huang, Xu, Zhang, and Zhang]{deepseekai2025deepseekr1incentivizingreasoningcapability}
DeepSeek-AI, Guo, D., Yang, D., Zhang, H., Song, J., Zhang, R., Xu, R., Zhu, Q., Ma, S., Wang, P., Bi, X., Zhang, X., Yu, X., Wu, Y., Wu, Z.~F., Gou, Z., Shao, Z., Li, Z., Gao, Z., Liu, A., Xue, B., Wang, B., Wu, B., Feng, B., Lu, C., Zhao, C., Deng, C., Zhang, C., Ruan, C., Dai, D., Chen, D., Ji, D., Li, E., Lin, F., Dai, F., Luo, F., Hao, G., Chen, G., Li, G., Zhang, H., Bao, H., Xu, H., Wang, H., Ding, H., Xin, H., Gao, H., Qu, H., Li, H., Guo, J., Li, J., Wang, J., Chen, J., Yuan, J., Qiu, J., Li, J., Cai, J.~L., Ni, J., Liang, J., Chen, J., Dong, K., Hu, K., Gao, K., Guan, K., Huang, K., Yu, K., Wang, L., Zhang, L., Zhao, L., Wang, L., Zhang, L., Xu, L., Xia, L., Zhang, M., Zhang, M., Tang, M., Li, M., Wang, M., Li, M., Tian, N., Huang, P., Zhang, P., Wang, Q., Chen, Q., Du, Q., Ge, R., Zhang, R., Pan, R., Wang, R., Chen, R.~J., Jin, R.~L., Chen, R., Lu, S., Zhou, S., Chen, S., Ye, S., Wang, S., Yu, S., Zhou, S., Pan, S., Li, S.~S., Zhou, S., Wu, S., Ye, S., Yun, T., Pei, T., Sun, T., Wang, T., Zeng, W.,
  Zhao, W., Liu, W., Liang, W., Gao, W., Yu, W., Zhang, W., Xiao, W.~L., An, W., Liu, X., Wang, X., Chen, X., Nie, X., Cheng, X., Liu, X., Xie, X., Liu, X., Yang, X., Li, X., Su, X., Lin, X., Li, X.~Q., Jin, X., Shen, X., Chen, X., Sun, X., Wang, X., Song, X., Zhou, X., Wang, X., Shan, X., Li, Y.~K., Wang, Y.~Q., Wei, Y.~X., Zhang, Y., Xu, Y., Li, Y., Zhao, Y., Sun, Y., Wang, Y., Yu, Y., Zhang, Y., Shi, Y., Xiong, Y., He, Y., Piao, Y., Wang, Y., Tan, Y., Ma, Y., Liu, Y., Guo, Y., Ou, Y., Wang, Y., Gong, Y., Zou, Y., He, Y., Xiong, Y., Luo, Y., You, Y., Liu, Y., Zhou, Y., Zhu, Y.~X., Xu, Y., Huang, Y., Li, Y., Zheng, Y., Zhu, Y., Ma, Y., Tang, Y., Zha, Y., Yan, Y., Ren, Z.~Z., Ren, Z., Sha, Z., Fu, Z., Xu, Z., Xie, Z., Zhang, Z., Hao, Z., Ma, Z., Yan, Z., Wu, Z., Gu, Z., Zhu, Z., Liu, Z., Li, Z., Xie, Z., Song, Z., Pan, Z., Huang, Z., Xu, Z., Zhang, Z., and Zhang, Z.
\newblock Deepseek-r1: Incentivizing reasoning capability in llms via reinforcement learning, 2025.
\newblock URL \url{https://arxiv.org/abs/2501.12948}.

\bibitem[Guha et~al.(2025)Guha, Marten, Keh, Raoof, Smyrnis, Bansal, Nezhurina, Mercat, Vu, Sprague, Suvarna, Feuer, Chen, Khan, Frankel, Grover, Choi, Muennighoff, Su, Zhao, Yang, Pimpalgaonkar, Sharma, Ji, Deng, Pratt, Ramanujan, Saad-Falcon, Li, Dave, Albalak, Arora, Wulfe, Hegde, Durrett, Oh, Bansal, Gabriel, Grover, Chang, Shankar, Gokaslan, Merrill, Hashimoto, Choi, Jitsev, Heckel, Sathiamoorthy, Dimakis, and Schmidt]{guha2025openthoughtsdatarecipesreasoning}
Guha, E., Marten, R., Keh, S., Raoof, N., Smyrnis, G., Bansal, H., Nezhurina, M., Mercat, J., Vu, T., Sprague, Z., Suvarna, A., Feuer, B., Chen, L., Khan, Z., Frankel, E., Grover, S., Choi, C., Muennighoff, N., Su, S., Zhao, W., Yang, J., Pimpalgaonkar, S., Sharma, K., Ji, C. C.-J., Deng, Y., Pratt, S., Ramanujan, V., Saad-Falcon, J., Li, J., Dave, A., Albalak, A., Arora, K., Wulfe, B., Hegde, C., Durrett, G., Oh, S., Bansal, M., Gabriel, S., Grover, A., Chang, K.-W., Shankar, V., Gokaslan, A., Merrill, M.~A., Hashimoto, T., Choi, Y., Jitsev, J., Heckel, R., Sathiamoorthy, M., Dimakis, A.~G., and Schmidt, L.
\newblock Openthoughts: Data recipes for reasoning models, 2025.
\newblock URL \url{https://arxiv.org/abs/2506.04178}.

\bibitem[He et~al.(2025)He, Liu, Liu, Yan, Wang, Cheng, Zhang, Zhang, Xu, Shen, Li, Zeng, Wei, Cheng, An, Liu, and Zhou]{he2025skywork}
He, J., Liu, J., Liu, C.~Y., Yan, R., Wang, C., Cheng, P., Zhang, X., Zhang, F., Xu, J., Shen, W., Li, S., Zeng, L., Wei, T., Cheng, C., An, B., Liu, Y., and Zhou, Y.
\newblock Skywork open reasoner 1 technical report.
\newblock \emph{arXiv preprint arXiv:2505.22312}, 2025.

\bibitem[Hendrycks et~al.(2021)Hendrycks, Burns, Kadavath, Arora, Basart, Tang, Song, and Steinhardt]{hendrycksmath2021}
Hendrycks, D., Burns, C., Kadavath, S., Arora, A., Basart, S., Tang, E., Song, D., and Steinhardt, J.
\newblock Measuring mathematical problem solving with the math dataset.
\newblock \emph{NeurIPS}, 2021.

\bibitem[Hu et~al.(2025)Hu, Zhang, Han, Jiang, Zhang, and Shum]{hu2025openreasonerzeroopensourceapproach}
Hu, J., Zhang, Y., Han, Q., Jiang, D., Zhang, X., and Shum, H.-Y.
\newblock Open-reasoner-zero: An open source approach to scaling up reinforcement learning on the base model, 2025.
\newblock URL \url{https://arxiv.org/abs/2503.24290}.

\bibitem[HuggingFace()]{Math-Verify}
HuggingFace.
\newblock Math-verify.
\newblock \url{https://github.com/huggingface/Math-Verify}.

\bibitem[Jones(2021)]{jones2021scalingscalinglawsboard}
Jones, A.~L.
\newblock Scaling scaling laws with board games, 2021.
\newblock URL \url{https://arxiv.org/abs/2104.03113}.

\bibitem[Jumper et~al.(2021)Jumper, Evans, Pritzel, Green, Figurnov, Ronneberger, Tunyasuvunakool, Bates, {\v Z}{\'\i}dek, Potapenko, Bridgland, Meyer, Kohl, Ballard, Cowie, Romera-Paredes, Nikolov, Jain, Adler, Back, Petersen, Reiman, Clancy, Zielinski, Steinegger, Pacholska, Berghammer, Bodenstein, Silver, Vinyals, Senior, Kavukcuoglu, Kohli, and Hassabis]{alphafold2021}
Jumper, J., Evans, R., Pritzel, A., Green, T., Figurnov, M., Ronneberger, O., Tunyasuvunakool, K., Bates, R., {\v Z}{\'\i}dek, A., Potapenko, A., Bridgland, A., Meyer, C., Kohl, S. A.~A., Ballard, A.~J., Cowie, A., Romera-Paredes, B., Nikolov, S., Jain, R., Adler, J., Back, T., Petersen, S., Reiman, D., Clancy, E., Zielinski, M., Steinegger, M., Pacholska, M., Berghammer, T., Bodenstein, S., Silver, D., Vinyals, O., Senior, A.~W., Kavukcuoglu, K., Kohli, P., and Hassabis, D.
\newblock Highly accurate protein structure prediction with alphafold.
\newblock \emph{Nature}, 596\penalty0 (7873):\penalty0 583--589, 2021.
\newblock \doi{10.1038/s41586-021-03819-2}.
\newblock URL \url{https://doi.org/10.1038/s41586-021-03819-2}.

\bibitem[Kestin et~al.(2024)Kestin, Miller, Klales, Milbourne, and Ponti]{Kestin__2024}
Kestin, G., Miller, K., Klales, A., Milbourne, T., and Ponti, G.
\newblock Ai tutoring outperforms active learning.
\newblock May 2024.
\newblock \doi{10.21203/rs.3.rs-4243877/v1}.
\newblock URL \url{http://dx.doi.org/10.21203/rs.3.rs-4243877/v1}.

\bibitem[LI et~al.(2024)LI, Beeching, Tunstall, Lipkin, Soletskyi, Huang, Rasul, Yu, Jiang, Shen, Qin, Dong, Zhou, Fleureau, Lample, and Polu]{numina_math_datasets}
LI, J., Beeching, E., Tunstall, L., Lipkin, B., Soletskyi, R., Huang, S.~C., Rasul, K., Yu, L., Jiang, A., Shen, Z., Qin, Z., Dong, B., Zhou, L., Fleureau, Y., Lample, G., and Polu, S.
\newblock Numinamath.
\newblock \url{[https://huggingface.co/AI-MO/NuminaMath-1.5](https://github.com/project-numina/aimo-progress-prize/blob/main/report/numina_dataset.pdf)}, 2024.

\bibitem[Li et~al.(2025)Li, Zou, and Liu]{limr2025}
Li, X., Zou, H., and Liu, P.
\newblock Limr: Less is more for rl scaling.
\newblock \url{https://github.com/GAIR-NLP/LIMR}, 2025.

\bibitem[Lin(1992)]{er-lin92}
Lin, L.-J.
\newblock Self-improving reactive agents based on reinforcement learning, planning and teaching.
\newblock \emph{Machine Learning}, 8\penalty0 (3):\penalty0 293--321, 1992.
\newblock \doi{10.1007/BF00992699}.
\newblock URL \url{https://doi.org/10.1007/BF00992699}.

\bibitem[Liu et~al.(2024)Liu, Zheng, Qiao, Duan, Fei, Zhou, Zhang, Zhang, Lin, and Chen]{liu2024mathbench}
Liu, H., Zheng, Z., Qiao, Y., Duan, H., Fei, Z., Zhou, F., Zhang, W., Zhang, S., Lin, D., and Chen, K.
\newblock Mathbench: Evaluating the theory and application proficiency of llms with a hierarchical mathematics benchmark.
\newblock In \emph{Findings of the Association for Computational Linguistics ACL 2024}, pp.\  6884--6915, 2024.

\bibitem[Liu et~al.(2025)Liu, Chen, Li, Qi, Pang, Du, Lee, and Lin]{liu2025understandingr1zeroliketrainingcritical}
Liu, Z., Chen, C., Li, W., Qi, P., Pang, T., Du, C., Lee, W.~S., and Lin, M.
\newblock Understanding r1-zero-like training: A critical perspective, 2025.
\newblock URL \url{https://arxiv.org/abs/2503.20783}.

\bibitem[Metropolis \& Ulam(1949)Metropolis and Ulam]{Metropolis:1949:MCM}
Metropolis, N. and Ulam, S.
\newblock The {Monte Carlo} method.
\newblock \emph{j-J-AM-STAT-ASSOC}, 44\penalty0 (247):\penalty0 335--341, September 1949.
\newblock ISSN 0162-1459 (print), 1537-274X (electronic).
\newblock \doi{https://doi.org/10.2307/2280232}.
\newblock URL \url{http://links.jstor.org/sici?sici=0162-1459%28194909%2944%3A247%3C335%3ATMCM%3E2.0.CO%3B2-3; http://www.jstor.org/stable/2280232}.

\bibitem[Muennighoff et~al.(2025)Muennighoff, Yang, Shi, Li, Fei-Fei, Hajishirzi, Zettlemoyer, Liang, Candès, and Hashimoto]{muennighoff2025s1simpletesttimescaling}
Muennighoff, N., Yang, Z., Shi, W., Li, X.~L., Fei-Fei, L., Hajishirzi, H., Zettlemoyer, L., Liang, P., Candès, E., and Hashimoto, T.
\newblock s1: Simple test-time scaling, 2025.
\newblock URL \url{https://arxiv.org/abs/2501.19393}.

\bibitem[NVIDIA()]{NeMo}
NVIDIA.
\newblock Nemo curator: The gpu-accelerated open source framework for efficient generative ai model data curation.
\newblock \url{https://github.com/NVIDIA/NeMo-Curator}.

\bibitem[OpenAI(2024)]{openai2024o1}
OpenAI.
\newblock Learning to reason with llms.
\newblock Open AI blog, 2024.
\newblock URL \url{https://openai.com/index/learning-to-reason-with-llms/}.

\bibitem[OpenAI(2025)]{openai2025o3-4}
OpenAI.
\newblock Introducing openai o3 and o4-mini.
\newblock Open AI blog, 2025.
\newblock URL \url{https://openai.com/index/introducing-o3-and-o4-mini/}.

\bibitem[Qin et~al.(2024)Qin, Li, Zou, Liu, Xia, Huang, Ye, Yuan, Liu, Li, and Liu]{o1journey}
Qin, Y., Li, X., Zou, H., Liu, Y., Xia, S., Huang, Z., Ye, Y., Yuan, W., Liu, Z., Li, Y., and Liu, P.
\newblock O1 replication journey: A strategic progress report – part 1.
\newblock \emph{arXiv preprint arXiv:2410.18982}, 2024.
\newblock URL \url{https://arxiv.org/abs/2410.18982}.

\bibitem[QwenTeam()]{QwQ-32B}
QwenTeam.
\newblock Qwq-32b: Embracing the power of reinforcement learning.
\newblock \url{https://qwenlm.github.io/blog/qwq-32b/}.

\bibitem[Schulman et~al.(2017)Schulman, Wolski, Dhariwal, Radford, and Klimov]{schulman2017proximal}
Schulman, J., Wolski, F., Dhariwal, P., Radford, A., and Klimov, O.
\newblock Proximal policy optimization algorithms.
\newblock \emph{arXiv preprint arXiv:1707.06347}, 2017.

\bibitem[Scott(2010)]{scott2010Bayesianbandit}
Scott, S.~L.
\newblock A modern bayesian look at the multi-armed bandit.
\newblock \emph{Applied Stochastic Models in Business and Industry}, 26\penalty0 (6):\penalty0 639--658, 2010.
\newblock \doi{https://doi.org/10.1002/asmb.874}.
\newblock URL \url{https://onlinelibrary.wiley.com/doi/abs/10.1002/asmb.874}.

\bibitem[Shao et~al.(2024)Shao, Wang, Zhu, Xu, Song, Bi, Zhang, Zhang, Li, Wu, and Guo]{shao2024deepseekmathpushinglimitsmathematical}
Shao, Z., Wang, P., Zhu, Q., Xu, R., Song, J., Bi, X., Zhang, H., Zhang, M., Li, Y.~K., Wu, Y., and Guo, D.
\newblock Deepseekmath: Pushing the limits of mathematical reasoning in open language models, 2024.
\newblock URL \url{https://arxiv.org/abs/2402.03300}.

\bibitem[Silver et~al.(2016)Silver, Huang, Maddison, Guez, Sifre, van~den Driessche, Schrittwieser, Antonoglou, Panneershelvam, Lanctot, Dieleman, Grewe, Nham, Kalchbrenner, Sutskever, Lillicrap, Leach, Kavukcuoglu, Graepel, and Hassabis]{alphago2016}
Silver, D., Huang, A., Maddison, C.~J., Guez, A., Sifre, L., van~den Driessche, G., Schrittwieser, J., Antonoglou, I., Panneershelvam, V., Lanctot, M., Dieleman, S., Grewe, D., Nham, J., Kalchbrenner, N., Sutskever, I., Lillicrap, T., Leach, M., Kavukcuoglu, K., Graepel, T., and Hassabis, D.
\newblock Mastering the game of go with deep neural networks and tree search.
\newblock \emph{Nature}, 529:\penalty0 484--503, 2016.
\newblock URL \url{http://www.nature.com/nature/journal/v529/n7587/full/nature16961.html}.

\bibitem[Snoswell et~al.(2020)Snoswell, Singh, and Ye]{snoswell2020revisiting}
Snoswell, A.~J., Singh, S.~P., and Ye, N.
\newblock Revisiting maximum entropy inverse reinforcement learning: New perspectives and algorithms.
\newblock In \emph{2020 IEEE Symposium Series on Computational Intelligence (SSCI)}, pp.\  241--249. IEEE, 2020.

\bibitem[Team(2025)]{kimiteam2025kimik15scalingreinforcement}
Team, K.
\newblock Kimi k1.5: Scaling reinforcement learning with llms, 2025.
\newblock URL \url{https://arxiv.org/abs/2501.12599}.

\bibitem[Tesauro(1994)]{10.1162/neco.1994.6.2.215}
Tesauro, G.
\newblock Td-gammon, a self-teaching backgammon program, achieves master-level play.
\newblock \emph{Neural Comput.}, 6\penalty0 (2):\penalty0 215–219, March 1994.
\newblock ISSN 0899-7667.
\newblock \doi{10.1162/neco.1994.6.2.215}.
\newblock URL \url{https://doi.org/10.1162/neco.1994.6.2.215}.

\bibitem[Wang \& Fan(2025)Wang and Fan]{wanggptlearning2025}
Wang, J. and Fan, W.
\newblock The effect of chatgpt on students'learning performance, learning perception, and higher-order thinking: insights from a meta-analysis.
\newblock \emph{Humanities and Social Sciences Communications}, 12\penalty0 (1):\penalty0 621, 2025.
\newblock \doi{10.1057/s41599-025-04787-y}.
\newblock URL \url{https://doi.org/10.1057/s41599-025-04787-y}.

\bibitem[Wang et~al.(2024)Wang, Li, Lian, Ma, Song, and Wei]{wang2024mitigatinglanguagemismatchrepetition}
Wang, W., Li, Z., Lian, D., Ma, C., Song, L., and Wei, Y.
\newblock Mitigating the language mismatch and repetition issues in llm-based machine translation via model editing, 2024.
\newblock URL \url{https://arxiv.org/abs/2410.07054}.

\bibitem[Wang et~al.(2025)Wang, Cui, Wan, and Zhao]{wang2025dumpautomateddistributionlevelcurriculum}
Wang, Z., Cui, G., Wan, K., and Zhao, W.
\newblock Dump: Automated distribution-level curriculum learning for rl-based llm post-training, 2025.
\newblock URL \url{https://arxiv.org/abs/2504.09710}.

\bibitem[Wei et~al.(2023)Wei, Luan, Liu, Dong, and Wang]{wei2023cmath}
Wei, T., Luan, J., Liu, W., Dong, S., and Wang, B.
\newblock Cmath: Can your language model pass chinese elementary school math test?, 2023.

\bibitem[Xue et~al.(2023)Xue, Fu, Zhou, Zheng, and You]{10.5555/3666122.3668712}
Xue, F., Fu, Y., Zhou, W., Zheng, Z., and You, Y.
\newblock To repeat or not to repeat: insights from scaling llm under token-crisis.
\newblock In \emph{Proceedings of the 37th International Conference on Neural Information Processing Systems}, NIPS '23, Red Hook, NY, USA, 2023. Curran Associates Inc.

\bibitem[Yang et~al.(2025)Yang, Li, Yang, Zhang, Hui, Zheng, Yu, Gao, Huang, Lv, et~al.]{yang2025qwen3}
Yang, A., Li, A., Yang, B., Zhang, B., Hui, B., Zheng, B., Yu, B., Gao, C., Huang, C., Lv, C., et~al.
\newblock Qwen3 technical report.
\newblock \emph{arXiv preprint arXiv:2505.09388}, 2025.

\bibitem[Ye et~al.(2025)Ye, Huang, Xiao, Chern, Xia, and Liu]{ye2025limoreasoning}
Ye, Y., Huang, Z., Xiao, Y., Chern, E., Xia, S., and Liu, P.
\newblock Limo: Less is more for reasoning, 2025.
\newblock URL \url{https://arxiv.org/abs/2502.03387}.

\bibitem[Yu et~al.(2025)Yu, Zhang, Zhu, Yuan, Zuo, Yue, Dai, Fan, Liu, Liu, Liu, Lin, Lin, Ma, Sheng, Tong, Zhang, Zhang, Zhang, Zhu, Zhu, Chen, Chen, Wang, Yu, Song, Wei, Zhou, Liu, Ma, Zhang, Yan, Qiao, Wu, and Wang]{yu2025dapoopensourcellmreinforcement}
Yu, Q., Zhang, Z., Zhu, R., Yuan, Y., Zuo, X., Yue, Y., Dai, W., Fan, T., Liu, G., Liu, L., Liu, X., Lin, H., Lin, Z., Ma, B., Sheng, G., Tong, Y., Zhang, C., Zhang, M., Zhang, W., Zhu, H., Zhu, J., Chen, J., Chen, J., Wang, C., Yu, H., Song, Y., Wei, X., Zhou, H., Liu, J., Ma, W.-Y., Zhang, Y.-Q., Yan, L., Qiao, M., Wu, Y., and Wang, M.
\newblock Dapo: An open-source llm reinforcement learning system at scale, 2025.
\newblock URL \url{https://arxiv.org/abs/2503.14476}.

\bibitem[Zhang et~al.(2023)Zhang, Li, Zong, Ying, He, and Qiu]{zhang2023evaluating}
Zhang, X., Li, C., Zong, Y., Ying, Z., He, L., and Qiu, X.
\newblock Evaluating the performance of large language models on gaokao benchmark.
\newblock \emph{arXiv preprint arXiv:2305.12474}, 2023.

\bibitem[Zhao et~al.(2024)Zhao, Yin, Zeng, Wang, Shi, Lyu, Wang, Luo, and Zhang]{zhao2024marcoo1openreasoningmodels}
Zhao, Y., Yin, H., Zeng, B., Wang, H., Shi, T., Lyu, C., Wang, L., Luo, W., and Zhang, K.
\newblock Marco-o1: Towards open reasoning models for open-ended solutions, 2024.
\newblock URL \url{https://arxiv.org/abs/2411.14405}.

\bibitem[Zhu et~al.(2024)Zhu, Zhang, and Wang]{zhu2024embracingaieducationunderstanding}
Zhu, T., Zhang, K., and Wang, W.~Y.
\newblock Embracing ai in education: Understanding the surge in large language model use by secondary students, 2024.
\newblock URL \url{https://arxiv.org/abs/2411.18708}.

\end{thebibliography}





\end{document}